\documentclass{article}
\pdfoutput=1

\PassOptionsToPackage{numbers}{natbib}

\usepackage[preprint]{neurips_2023}

\usepackage[utf8]{inputenc} 
\usepackage[T1]{fontenc}    
\usepackage{url}            
\usepackage{booktabs}       
\usepackage{amsfonts}       
\usepackage{nicefrac}       
\usepackage{microtype}      
\usepackage{xcolor}         
\usepackage{amsmath}        
\usepackage{amssymb}
\usepackage{multirow}
\usepackage[ruled,vlined]{algorithm2e}
\usepackage{url}
\usepackage{calc}
\usepackage{lettrine}
\usepackage{graphicx}
\usepackage{orcidlink}
\usepackage{hyperref}       
\title{Global Layers: Non-IID Tabular Federated Learning}

\author{%
  Yazan ~Obeidi \orcidlink{0000-0002-4228-9610} \\
  IBM \\
  \texttt{yazan.obeidi@ibm.com} \\
}

\begin{document}
\hypersetup{pdfauthor={Yazan Obeidi},pdftitle={Global Layers: Non-IID Tabular Federated Learning}}

\maketitle

\begin{abstract}
Data heterogeneity between clients remains a key challenge in Federated Learning (FL), particularly in the case of tabular data.
This work presents Global Layers (GL), a novel partial model personalization method robust in the presence of joint distribution $P(X,Y)$ shift and mixed input/output spaces $X \times Y$ across clients.
To the best of our knowledge, GL is the first method capable of supporting both client-exclusive features and classes.
We introduce two new benchmark experiments for tabular FL naturally partitioned from existing real world datasets: i) UCI Covertype split into 4 clients by "wilderness area" feature, and ii) UCI Heart Disease, SAHeart, UCI Heart Failure, each as clients.
Empirical results in these experiments in the full-participant setting show that GL achieves better outcomes than Federated Averaging (FedAvg) and local-only training, with some clients even performing better than their centralized baseline. 
\end{abstract}

\section{Introduction}
\label{introduction}

Federated Learning (FL) is an emerging framework for collaboratively training machine learning models across distributed private datasets, termed clients.
First proposed as a way to efficiently train language models over large numbers of edge devices \citep{mcmahan2017}, FL has been since recognized more generally as a privacy-preserving alternative to centralized learning \citep{kairouz21}.
This has attracted recent interest in enterprise applications, as centralizing data may be prohibited by regulation or may have unwanted business impact such as trade secret disclosure.

A prominent scenario is training diagnostic models across hospitals \citep{prayitno21, sheller20, chen21}.
A hospital may have insufficient data to train a high quality model alone.
FL enables model training while data remains at source.
When the assumption that the $K$ clients are independent and identically distributed (IID) samples from a single meta-distribution is satisfied, i.e. $k \overset{iid}\sim P(X,Y|k) \,\forall k \in K$, we can expect the resulting federated model to converge with performance comparable to a centralized model \citep{haddadpour19}.

Unfortunately, this assumption is rarely met in practice.
Geographically distributed data sources will inherently reflect their distributional biases.
Consider that hospitals may address disparate patient demographics $P(X)$ with distinct disease susceptibilities $P(X|Y)$, treatment responses $P(Y|X)$, and outcomes $P(Y)$ \citep{rubin1974, szczepura05, zhong22}.
Distribution shift among clients causes their local gradient estimates to drift \citep{karimireddy21} resulting in reduced or stalled global model convergence \citep{haddadpour19, zhao18, zhu21, li20, mitra21}.
Moreover, hospitals may maintain different medical record fields for their $X$ and $Y$ due to varying interventions, testing equipment, encoding choices, and regulatory requirements.
Yet, heterogeneous input/output spaces are incompatible with full-parameter federated models.

Many early FL works tried to avoid the non-IID issue entirely by focusing on special cases.
Horizontal FL methods assume the existence of a sufficiently explanatory in-distribution subset of "common grounds" features and consistent target spaces \citep{kairouz21}.
Vertical FL methods support heterogeneous features spaces but assume a sufficiently large alignment in sample space \citep{liu22}.
Existing transfer FL methods support missing features and labels among clients and assume overlapping sample and feature space. 
In all three cases, conditional distribution shift and target space inconsistency remain as challenges, and we lose all client specific information that could have been beneficial for learning.
None of these settings can address our hospital scenario, and others scenarios that place no assumptions of overlapping feature or samples and where the joint distribution $P(X,Y)$ and spaces $X \times Y$ may arbitrarily differ across clients $k \in K$.

To the best of our knowledge, no method so far supports simultaneous feature and label space heterogeneity across clients without assumptions of overlapping features, labels, or samples.
We could only find two works that propose and study methods supporting disjoint features but not labels \citep{liu_completely_2022, rakotomamonjy23} and vice-versa \citep{makhija22, zhang23}.
Second, while tabular data format is the most common in business \citep{borisov22}, the majority of works
study only image and text datasets.
These already have a natural shared representation space across tasks  e.g. pixels and token sequences, often also preserving their labels, with a rich spatial or temporal structure, unlike tabular data, which contains arbitrarily ordered ordinal and numeric features.

In this work, we propose Global Layers (GL), a novel partial model personalization
framework capable of addressing the most general case of statistical heterogeneity in FL.
GL strongly separates private client input/output layers and global inner federated layers with the goal of learning personalized compositions of global equipredictive transformations.
GL places no restriction on the term that varies in the joint distribution $P(X,Y)$ across clients and makes no assumptions of overlap or specific ordering in either the features, targets or samples across clients.

To empirically analyze GL and given a lack of existing real-world non-IID tabular FL benchmark experiments appropriate for our setting, we introduce two new experiments constructed from existing real-world public datasets 
with natural data partitions: i) UCI Covertype split by "wilderness area" feature into 4 regionally-distributed, single-source clients, and ii) 3 independent heart disease classification datasets with each as a client: UCI Heart Disease, SAHeart, and UCI Heart Failure, collected from hospitals in USA, South Africa, and Pakistan, respectively.

Our results in both experiments in the full-participant setting show that GL achieves better outcomes than both full-parameter aggregation with Federated Averaging (FedAvg) and local-only training, with some clients even performing better than their centralized baseline.

\section{Background and Related Work}
\label{background}

\subsection{Federated Learning}

The standard FL objective is one global model $w \in \mathbb{R}^d$ that minimizes the aggregated local losses:
\begin{equation} 
\label{eq:fedavg}
\underset{w} \min\;F(w) := \mathbb{E}_{k}[F_{k}(w)] = \sum_{k}^{K}{\alpha_{k}F_{k}(w)}
\end{equation} 
Where the loss for client $k \in K$ with private labeled data $z_k = (x_k, y_k)$ is:
\begin{equation}
\label{eq:local-fedavg}
F_{k}(w)=\mathbb{E}_{z_{k} \sim P_{k}}[f_{k}(w)] =\frac{1}{n_{k}}\sum_{i=1}^{n_{k}}f_{k}(w, z_{k,i})
\end{equation}
FedAvg \citep{mcmahan2017} takes a weighted average for $a_{k}$, where $n_{k}$ is the number of samples in the $k$th client:
\begin{equation} 
\label{eq:fedavgP}
\alpha_{k} = \frac{n_{k}}{  \sum_{k}^{K}{n_{k}}}
\end{equation}
Implicit in this form is the assumption of an aggregated target distribution \citep{mohri19}:
\begin{equation}
\label{eq:uniform}
P = \sum_{k}^{K} \alpha_{k} P_{k}
\end{equation}
For a moment let us set aside the dimensionality implications of using the same $w$ for all clients.
The assumption in \ref{eq:uniform} illustrates
why full-parameter federation with a single global model $w$ struggles in the presence of distribution shift, and in particular concept drift, among clients.
A single global model effectively interpolates between client distributions, giving each client the same averaged prediction for a particular input.
While we are instead interested in a model that extrapolates beyond client distributions and with client-appropriate predictions.

Consider a toy example of 2 clients $i, j$ performing binary classification where one conditional probability distribution is the compliment of the other, i.e. $P_i=1-P_j$. 
From \ref{eq:uniform} we see the learning objective becomes a uniform distribution.
A single global model will give predictions that are wrong for both clients, and will not converge as its gradients updates will tend to cancel out.

This limitation is applicable to any non-IID FL method relying on the "one model fits all" assumption \citep{mohri19, mansour20, criado22}.
For example, methods that attempt to address distribution shift through improvements in the optimization \citep{karimireddy21, li20, mohri19, deng21, li_feddane_20, reddi21, karimireddy_mime_2021, acar21, zhang_fedpd_21} can help with covariate and prior probability shift but they alone cannot address concept shift or concept drift.

\subsection{Personalized Federated Learning}

\paragraph{Full-model personalization} These methods address concept shift and concept drift by permitting local model divergence from the global model $w$.
While the assumption of consistent input/output dimensionality remains, we again set this aside for a moment and focus only on distribution shift.

Fine-tuning \citep{yu22, wang19, mansour20} can help with concept shift and concept drift \citep{kairouz21} but fine-tuned models are vulnerable to local silo \citep{chen22} or global model \citep{albuquerque20} drift, and can over-fit, corrupt global information, and struggle to generalize OOD \citep{kumar22}. 
Pre-training \citep{nguyen22} faces similar issues.
Some works \citep{jiang19, chen19, khodak19, fallah20, zhang21} have explored meta-learning
to optimize for a well-initialized global model that best optimizes during local fine-tuning, observing improvement on these issues.
Though, these methods must calculate the Hessian, an expensive step, or approximate it with a performance loss \citep{dinh22}.

While fine-tuning methods operate in distinct phases, proximal operator methods \citep{dinh22, smith17, li21, huang21, mendieta022, anon23fedpac, dinh_fedu_22, guo22, banerjee22} simultaneously train separate local-only and global parameters with a proximal penalty as a selective multi-task transfer learning mechanism.
These generally extend \ref{eq:fedavg} with penalty $R_{k}$, where $\Omega$ often contains some relevant global information, and $R_{k} = \frac{\lambda}{2} {\vert w_{k} - w \vert }^{2}$ is a common choice to penalize parameter divergence between local and global models.
\begin{equation} 
\label{eq:proximal}
\underset{w, \{w_{k}\}_{k=1}^{K}} \min\;\sum_{k}^{K}{\alpha_{k} \left( F_{k}(w_{k}) + R_{k}(w_{k}, \Omega) \right)}
\end{equation}
Related are interpolation methods that learn client-specific mixtures of local and global models, often also using a proximal penalty \citep{hanzely20, hanzely21} or a different mechanism \citep{mansour20, peterson19, deng20, reisser21, guo20, zhang_fomo_21, marfoq_fedem_22}.

Despite increased non-IID robustness over FedAvg in empirical and theoretical analyses, both proximal and interpolation methods restrict the extent that local models can drift from the global model, ultimately limiting local convergence potential and thus the extent of distribution shift they can address \citep{mendieta022}.
Consider the toy example from earlier; these methods would still be unable to address this case.
Though some works explore penalties $R_k$ independent of a global 
$w$, for example aligning on local model output divergence \citep{makhija22, zhang_parameterized_2021, tan22}, we lose any transfer learning benefits we could have otherwise obtained in the presence of distribution shift.
In general, designing $R_{k}$ is non-trivial as task relationships may not be well specified, particularly within latent space.

\paragraph{Partial-model personalization}
The methods we described for full-model personalization can all be extended to work with restricted parameters, for example fine-tuning \citep{pillutla2022}, meta-learning \citep{zhang21}, and proximal methods \citep{chen21, zhou22}.
A general form is that $w_{k} = (u, v_{k})$ is the full model with global $u$ and local $v_{k}$ components and \ref{eq:fedavg} extends to:
\begin{equation} 
\label{eq:partial}
\underset{w, \{w_{k}\}_{k=1}^{K}} \min\;\sum_{k}^{K}{\alpha_{k} F_{k}( w_{k}) }
\end{equation}
While federating a subset of parameters allows for flexibility in input/output dimensionality and capacity in the other private parameters $v_{k}$, we mentioned earlier that only recently have a handful of works begun to analyze cases of non-overlapping features \citep{liu_completely_2022, rakotomamonjy23}, and labels \citep{makhija22, zhang23}, but not yet both.
Other works so far have assumed consistent feature and label spaces and restricted their focus on distribution shift.
For example, LG-FedAvg \citep{liang20} proposed personalized base layers to extract invariant representations, yet studies only covariate shift caused by 90 degree image rotations.
Similarly, FedPer \citep{arivazhagan19} and other works \citep{zhong22, tan22, collins21, chen_bridging_22} propose a global base with personalized heads but again make specific assumptions of distribution consistency and focus on image and text.
A few works have combined other methods with personalized layers such as partial meta-learning in PMFL \citep{zhang21}, proximal methods \citep{chen21, zhou22}, and more \citep{jiang22, singhal22}.

\subsection{Out-Of-Distribution Learning}

Learning in the presence of distribution shift has an extensive history of study in the centralized setting \citep{blanchard11, muandet13, storkey_2013, arjovsky21}, where the out-of-distribution (OOD) problem refers to training a predictor on limited coverage training supports that can generalize over unobserved test distributions \citep{koyama_when_2021}.
Remaining consistent with notation used in OOD literature, a common treatment \citep{wu_handling_2022} is that there exists a potentially unknown environmental variable $e \in \mathcal{E}$ conditioning datasets sampled from a meta-distribution $P(X, Y|e)=P(X|e)P(Y|X,e)$ where $\mathcal{E}_{train} \subseteq \mathcal{E}$ is the training data which can be multi-source and $Supp(\mathcal{E})$ is the space of all possible test environments.
This is a problematic scenario for Empirical Risk Minimization \citep{vapnik1991} (ERM) whose asymptotically optimal learning principle depends on sufficient in-distribution training coverage.

A recent line of works \citep{arjovsky21, koyama_when_2021, wu_handling_2022, gulrajani20, khezeli21, krueger21} have proposed alternatives to ERM that attempt to learn \textit{invariant predictors} that are able to generalize OOD while ignoring spurious correlations.
The key idea in all these works is that learning an \textit{equipredictive} representation  $\varphi (X)$ promotes learning invariant relationships across environments, described as $P(Y | \varphi (X), e) \,\forall e \in \mathcal{E}$.

A few recent works have explored applying these ideas to the FL setting \citep{chen_bridging_22, francis21, tang22, de_luca22} but have focused the scenario of generalizing from $k \in K$ training clients to a new unseen $k+1$ client.

\section{Global Layers (GL)}

\subsection{Methodology}

In this work we focus only on the full-participant setting and the generalization performance on each client's own cross-validation test set.
We extend the standard FL formulation to capture the type of data heterogeneity we mentioned earlier and that we present in our empirical analyses.

Each client dataset $D_k = \{ (x_{k,1}, y_{k,1}) ... (x_{k,n_k}, y_{k,n_k}) \} \in (X_k, Y_k)^{n_k}$ of size $n_k$ has a possibly disjoint feature space $X_k \in \mathbb{R}^{d_k}$ and target space $Y_k \in \mathbb{R}^{c_{k}}$, where $d_k$ is the number of local features, and $c_k$ is the number of local classes, with $c=1$ for regression.
Client datasets are IID samples $D_k \overset{iid} \sim P_k(X_k, Y_k)$ from different meta-distributions $P_k \overset{non-iid} \sim P(X, Y | k) \,\forall k \in K$. 

While our contribution does not require overlapping features, targets or samples, which is our goal, any existence of these would improve learning outcomes.
In practical terms, it would make sense for clients to have something in common, otherwise the reason for training together is unclear.

The key idea in our work is that rather than try to learn invariant relationships between client features and targets, which does not have clear definition when feature and target spaces are disjoint, is to instead try to learn invariant relationships within inner layer latent-space transformations across clients $P(Z_{L} | \varphi (Z_{L-1}), k) \,\forall k \in K$.
In particular, we want to learn these as personalized compositions with client specific dynamics.
The intuition is that despite distinct input/output spaces, we still wish to preserve shared invariant transformations.
At the same time, we do not want to restrict to the case where these relationships such as conditional distribution are fully preserved across clients.

\subsection{Learning Objective}
\label{learning-objective}

Inspired by the recent OOD literature, we propose learning a global equipredictive $\omega: H \rightarrow \hat{H}$ such that: 
$P(Y_k | v_{k} ( \omega (u_{k} (X_k))) \,\forall k \in K$, where $u_{k}: X_k \rightarrow H$ and $v_{k}: \hat{H} \rightarrow \hat{Y}_k$ are client specific input and output transformations, respectively.
We extend \ref{eq:partial} with $w_k = \{u_k, \omega, v_k\}$.
\begin{equation} 
\label{eq:gl}
\underset{w, \{u_{k}, v_{k}\}_{k=1}^{K}} \min\;F(u_k,\omega,v_k) := \sum_{k}^{K}{F_{k}(u_k, \omega, v_k)}
\end{equation}
Note that $a_{k}=1$ as we ensure consistent local gradient updates, elaborated in Section \ref{training-algorithm}.

The empirical loss at private sample $z_{k,i} \overset{iid}{\sim} P(X_k, Y_k)$ for the $k$th client is therefore:
\begin{equation} 
\label{eq:gl-sample}
f_{k, z_{k,i}}(w_k)
\end{equation}
Optimizing \ref{eq:gl} can be done following standard horizontal FL practices, including combining this objective with other personalization methods such as adding a proximal penalty term.

\subsection{Model Architecture}
\label{model-architecture}

The learning objective we just provided is architecturally agnostic, requiring only that $w_k$ can be composed into a feedforward model $f_{k}$ for each client that supports its local input/output dimensionality.
Generally the dimensionality, number of instances and task complexity will vary between clients which can motivate different capacities in private components.
Though we did explore this, in our experiments we found that consistent architectures across clients works well.
This leaves us with three main design choices for our overall model architecture: the choice of input layers $u$, output layers $v$ and inner global layers $\omega$.
We show this in Figure \ref{image:concept}.

In this work we only explored neural-network architectures for each, though in theory this is not a constraint.
Neural-networks have a well-studied ability to compose abstract and complex hierarchical features that can transfer across latent spaces \citep{yosinski_how_2014}, even in the tabular domain \citep{levin_transfer_2022}.

Recently some works have proposed Transformer \citep{vaswani17} based models for the tabular domain, such as TabTransformer \citep{huang21} and other similar architectures \citep{gorishniy21, luo20, zhang_iisan_21, somepalli2021, xie21} to capture high-order feature interactions across different subspaces, scales and even samples.
Such models have demonstrated to be competitive with even Gradient Decision Boosted Tree (GDBT) ensembles \citep{gorishniy21, gorishniy22} which was previously undisputed in its performance with tabular data.

In both of our experiments we use a TabTransformer inspired architecture with 6 stacked Transformer layers followed by a partially federated Multi-Layer Perceptron (MLP), shown in Figures \ref{image:model} and \ref{image:submodel}.

We first Batch Normalize \citep{ioffe15}, then pass all client features, including ordinal and numerical, through a private Entity Embedding-like \citep{guo16} layer.
Specifically, all features $x_k := \{x_{k,1}, ..., x_{k,d_k}\}$ receive their own fully-connected linear dense representation $\mathbf{e}$ parameterized by $\phi_k$:
\begin{equation} 
\label{eq:ee}
\mathbf{E}(x_k) := \{\mathbf{e}_{{\phi}_{k,1}}(x_{k,1}), ..., \mathbf{e}_{{\phi}_{k,d_k}}(x_{k,d_k})\}
\end{equation}
A dense layer per feature eliminates token conflicts.
In particular this is a mechanism for the model to learn to transform features independently to best embed local activations into the global latent space to mitigate both covariate shift and concept shift.
Intuitively, dissimilar embeddings can be encouraged to be similar when they share conditional likelihoods, and vice-versa.

Next are the 6 Transformer layers, after which we flatten and pass through the MLP.
The MLP is 5 feedforward layers alternating between a) a fully-connected layer with batchnorm, Scaled Exponential Linear Unit \citep{klambauer17} (SELU) activation and dropout, and b) a similar fully-connected layer but with self-gating and a residual skip connection. 
Only the second, third, and fourth MLP layers are federated, allowing for heterogeneous input/output dimensionality.
Decoupling client output layers also allows client models to mitigate concept drift.

See Appendix \ref{appendix:model} for much more details.

\subsection{Training Algorithm}
\label{training-algorithm}

Our proposed training algorithm \textit{BatchAlignedFedAvg} is based on standard FedAvg with two differences.
The first difference is that rather than specifying a batch size, instead we specify a total number of batches for each local epoch as a hyperparameter, which all clients use, giving us a consistent number of local gradient updates before the aggregation step and allowing us to set $a_{k}=1$.
This ensures each client contribution to the global update is proportional to its local empirical loss only, rather than the quantity of instances.
This also avoids local under or oversampling which can distort client distribution, particularly when sample sizes are low or imbalanced between clients.
The second difference is aggregation is done each time all clients complete a local pass on their batch.

Although in Algorithm \ref{BatchAlignedFedAvg} we show a local training step in the form of an SGD update, any optimizer can work, such as AdamW \citep{loshchilov_decoupled_2019}.

\begin{algorithm}
\SetAlgoLined
\SetKwInOut{Input}{Input}\SetKwInOut{Output}{output}
\Input{Initialized $\omega^{(0)}$, $\{u_k^{(0)}, v_k^{(0)}\}_{k=1}^{K}$, number of rounds $T$, clients $K$, learning rate $\gamma$, number of batches $B_n$, loss functions $F_k$ and global layer update rate $\eta$}

\For {$t=0,1, ..., T-1$} {
    Server broadcasts global layers $\omega^{(t)}$ to all clients\\
    \For {$b=0,1, ..., B_n-1$} {
        \For {clients $k \in K$ in parallel} {
        client trains locally \\
        $\{u_k, \omega_k, v_k\}^{(t+1)} = \{u_k, \omega_k, v_k\}^{(t)}-\gamma\hat{\nabla}F_k(u_k^{(t)}, \omega_k^{(t)}, v_k^{(t)})$ \\
        client sends updated global layers $\omega_k^{(t+1)}$ back to server
        }
        server aggregates all global layers $\omega^{(t+1)} = \sum_{k}^{K}  \omega_k^{(t+1)}$ \\
        \For {clients $k \in K$ in parallel} {
        client updates its global layers $\omega_k^{(t+1)} = \eta \omega_k^{(t+1)} + (1-\eta) \omega^{(t+1)}$
        }
    }
} 
\caption{\textit{BatchAlignedFedAvg}}
\label{BatchAlignedFedAvg}
\end{algorithm}

\section{Experiments}
\label{experiments}

We now describe the experiments we conducted to obtain the empirical results that justify our proposed method.

\subsection{Experiment Settings}
\label{experiment-settings}

We could not find existing suitable FL benchmark datasets that contained mixed input/output spaces, as such cases have only recently been motivated in non-IID FL settings.
We were cautious and chose to avoid randomly partitioning existing centralized benchmarks dataset, as extracting random subsets does not change the conditional distribution much, and conditioning on arbitrary attributes can introduce spurious correlations and destroy any existing real-world invariant relationships \citep{arjovsky21}.
Instead we took two approaches to construct realistic experiments with natural heterogeneity from existing public benchmark datasets.
These are shown in Table \ref{tab:datasets}, with original sources in Table \ref{tab:dataset-sources}.

In the first case we undid the multi-source centralization that was done in the Covertype dataset by separating 4 clients based on their "wilderness area" ordinal attribute.
This geographically locates each client to a slightly different region within Roosevelt National Forest of northern Colorado, with different elevations, species, and other local ecological attributes.
Each client has a different set of soil type classes, though incidentally there is some overlap.

In the second case we took 3 different datasets that classify heart disease: a) UCI Heart Disease \citep{detrano_international_1989} in the Cleveland, USA setting, b) South Africa Heart Disease \citep{chicco20}, and c) UCI Heart Failure from Faisalabad, Pakistan \citep{ahmad17}. 
These represent independent diagnostic factors leading to shared binary outcomes: no-presence or presence.
The feature spaces across these datasets are disjoint.

Both experiments use the same model with the only difference being larger embedding sizes for the larger Covertype datasets.
Models are randomly initialized with consistent random seeds across clients.

A fully detailed account on the construction of both experiments is provided in Appendix \ref{appendix:datasets}.

\subsection{Baseline Methods}
\label{baseline-methods}

In order to compare our numerical results obtained with GL we consider three baseline methods.
\begin{enumerate}
    \item\textbf{FedAvg} denotes a single global model $w$ where all parameters are aggregated during federation. As this is inconsistent with feature and target space differences across clients, we take the strategy of padding inconsistent dimensionality with 0's i.e. missing values, so that the spaces do align. We explain this more in the Appendix.
    \item\textbf{Local} denotes no federated aggregation done at all. In this case clients do separate local-only training over their private data.
    \item\textbf{Centralized} denotes breaking privacy and combining all data into a single centralized dataset. 
\end{enumerate}

\begin{table*}[t]
\begin{tabular}{ c c | c c | c c c c } 
\toprule
 Experiment & Dataset & Features & Classes & Train & Test & Validation & Total \\
\hline
\multirow{4}{*}{Covertype} 
& Comanche & 11 & 6  & 4772 & 247015 & 1578 & 253365  \\
& Neota & 11 & 3 & 385 & 29385 & 114 & 29884  \\
& Poudre & 11 & 4 & 3509 & 32293 & 1167 & 36969 \\
& Rawah & 11 & 4 & 2675 & 257199 & 922 & 260796  \\
& Centralized & 11 & 6 & 11340 & 565892 & 3780 & 581012  \\
\hline
\multirow{3}{*}{Heart Disease} 
& Cleveland & 13 & 2 & 182 & 100 & 21 & 304  \\
& South Africa & 9 & 2 & 278 & 153 & 31 & 462  \\
& Faisalabad & 12 & 2 & 180 & 100 & 20 & 299  \\
\hline
\end{tabular}
\centering
\caption{Summary of experiments and tabular datasets used}
\label{tab:datasets}
\end{table*}

\begin{table*}[t]
\begin{tabular}{ c c | c c c | c c c } 
\toprule
\multicolumn{2}{c}{} & \multicolumn{3}{|c|}{AUROC (\%)} & \multicolumn{3}{|c}{(Balanced) Accuracy$^1$ (\%)} \\
\hline
 Experiment & Dataset & Local & FedAvg & \textbf{GL} & Local & FedAvg & \textbf{GL} \\
\hline
\multirow{3}{*}{Covertype} 
& Comanche & $88.29$ & $79.58$ & $\mathbf{92.09}$ & $80.82$ & $80.46$ & $\mathbf{83.81}$  \\
& Neota & $79.44$ & $60.92$ & $\mathbf{88.85}$ & $76.43$ & $72.26$ & $\mathbf{81.13}$ \\
& Poudre & $86.74$ & $63.89$ & $\mathbf{88.88}$ & $72.57$ & $72.72$ & $\mathbf{74.21}$ \\
& Rawah  & $88.86$ & $69.82$ & $\mathbf{91.43}$ & $84.25$ & $83.65$ & $\mathbf{85.13}$ \\
& Centralized  & $88.93$ & $-$ & $-$ & $79.70$ & $-$ & $-$ \\
\hline
\multirow{3}{*}{Heart Disease} 
& Cleveland & $88.57$ & $\mathbf{89.67}$ & $89.59$ & $80.23$ & $78.06$ & $\mathbf{81.18}$  \\
& South Africa & $74.49$ & $71.13$ & $\mathbf{76.02}$ & $62.43$ & $56.15$ & $\mathbf{63.22}$ \\
& Faisalabad & $85.64$ & $83.68$ & $\mathbf{86.77}$ & $74.57$ & $67.66$ & $\mathbf{75.15}$ \\
\hline
\end{tabular}
\centering
\caption{Summary of experimental results. Mean cross-validation result reported across 21, 101 random seeds for Covertype, Heart Disease respectively. Standard deviation and 95\% confidence intervals provided in the Appendix. 
($^1$) Heart Disease binary classification reports balanced accuracy.}
\label{tab:numerical-results}
\end{table*}

\subsection{Numerical Results}
\label{numerical-results}

Table \ref{tab:numerical-results} summarizes our main results with two metrics: Area Under the Receiver Operating Characteristic Curve (AUROC) and accuracy; multi-class for Covertype, and balanced binary accuracy for Heart Disease.
We perform 21 and 101 cross-validation runs for Covertype and Heart Disease respectively. 
Although shown is the mean result, standard deviation and 95\% confidence interval are provided in Tables \ref{tab:numerical-results-auroc} and \ref{tab:numerical-results-acc} in Appendix \ref{appendix:numerical-results}.
We also provide an additional metric: Area Under the Precision Recall Curve (AUPRC) in Table \ref{tab:numerical-results-auprc}.

GL achieves significant lift with lower variance for all clients and metrics.
The single exception is the Cleveland client in Heart Disease where AUROC is relative within $0.09\%$, though for the same case, GL reports a $3.8\%$ relative lift for balanced binary accuracy.

For AUROC, GL reports on average about $4\%$ absolute lift for Covertype compared to the next highest scores between Local and FedAvg baselines, and about $1\%$ for Heart Disease.
Similarly, for (balanced) accuracy, GL lift is on average about $3\%$ and $1\%$ respectively compared to the next highest score.
For AUPRC, the average absolute lift is larger at about $7.5\%$ and $1.7\%$, respectively.
Compared to the worst performing baseline, GL reports absolute lift on average of $21.5\%$ and $3\%$ for AUROC, $4\%$ and $5.3\%$ for (balanced) accuracy, and $40\%$ and $4.3\%$ for AUPRC, for Covertype and Heart Disease respectively.

\subsection{Reproduction}
\label{reproduction}

Experiments were implemented in Torch 2.1.0 and Python 3.11.2.

In Appendix \ref{reproducibility} we describe how to reproduce our results.
Links to datasets are provided in Tables \ref{tab:dataset-sources} and \ref{tab:our-dataset-urls}.
Hyperparameters, including random seeds, are provided in Appendix \ref{appendix:hyperparameters}.
Experimental source code is available at: \url{https://github.com/transferFL/gl/}.
A README.md is provided in the repository with launch commands for each experiments that will replicate our numerical results.

With an Apple M2 MacBook Pro and CPU-only training, including repeated cross-validation runs, it takes about 3 hours for Covertype over 20 random seeds, and 2.5 hours for Heart Disease over 101 random seeds.

\section{Limitations and Future Work}

In this work we do not compare against other model personalization methods.
This remains as an important area for future work.

We also only study two experimental settings, mainly due to the lack of existing benchmark datasets appropriate for our study.
Conducting additional empirical analyses is also important future work.

Finally we have introduced a conceptual mechanism for learning in the presence of strong statistical heterogeneity between clients but we did not theoretically analyze this in-depth to provide convergence guarantees for the setting we introduce.
Having obtained these preliminary empirical results, future theoretical analyses have a clear motivation.

\section{Broader Impacts}

Privacy is a highly relevant social matter, particularly for organizations and their tabular data records.
Privacy-preserving methods that show better performance outcomes than their centralized counterparts such as ours create a natural incentive for their use that may promote further adoption and study.
This work takes an initial step forward in this direction.

\section{Conclusion}

We have introduced the first FL framework capable of supporting joint distribution shift and simultaneous feature and target space heterogeneity across clients with no assumptions of overlap in features, targets, or samples across clients.
We have presented empirical results on two real-world experiments with natural client partitions that both demonstrate better outcomes than standard FedAvg and local-only training, with some clients outperforming a centralized model.

\medskip

{\footnotesize
\bibliography{neurips_2023.bib}}

\newpage
\appendix
\section{Appendix}

This appendix section is structured as follows.
In Appendix \ref{appendix:non-iid} we review 4 cases of distributional shift with examples to develop intuitions.
In Appendix \ref{appendix:model}, we elaborate on the GL model architecture used in our experiments. In particular we describe which layers are federated, and thus are Global Layers, and which ones are the private layers.
In Appendix \ref{appendix:datasets}, we describe the construction of the datasets in our two experiments: i) Covertype with 4 client datasets, and ii) Heart Disease with 3 client datasets, all with cross-validation splits.
In Appendix \ref{appendix:experiments}, we provide additional details in each experimental setting, show additional numerical results that report standard deviation and 95\% confidence interval for AUROC, (Balanced) Accuracy, and AUPRC, and review the steps required for their reproducibility.

\section{Distribution Shift}
\label{appendix:non-iid}

Distributional shift occurs when the joint distribution $P(X,Y)$ varies across clients.
By considering each of its two expressions $P(Y|X)P(X)$ and $P(X|Y)P(Y)$ separately and varying one term while the other remains fixed, \citep{kairouz21} characterizes 4 types of distributional shift.
We briefly illustrate each with the scenario of a disease classifier $P(Y|X)$ and patient medical records $P(X)$ from hospitals belonging to a number of different countries.
Note that when one joint distribution term varies while the other is assumed fixed, one or both terms in the other expression must necessarily vary and is implied as such.

\begin{enumerate}
    \item \textbf{Covariate shift} occurs when the observed demographic $P(X)$ varies across hospitals while the disease outcome $P(Y|X)$ remain consistent. 
    \item \textbf{Prior probability shift} occurs when hospitals have similar demographics for the same outcomes $P(X|Y)$ yet have varying overall disease outcomes $P(Y)$.
    \item \textbf{Concept shift} occurs when hospitals share overall disease outcomes $P(Y)$ yet the demographics they observe for these $P(X|Y)$ vary.
    \item \textbf{Concept drift} occurs when the diagnostic criteria $P(Y|X)$ qualifying a disease outcome for a demographic $P(X)$ vary between hospitals.
\end{enumerate}

\section{Model}
\label{appendix:model}

\subsection{Architecture}

\begin{figure}[h]
\centering
\includegraphics[scale=0.9]{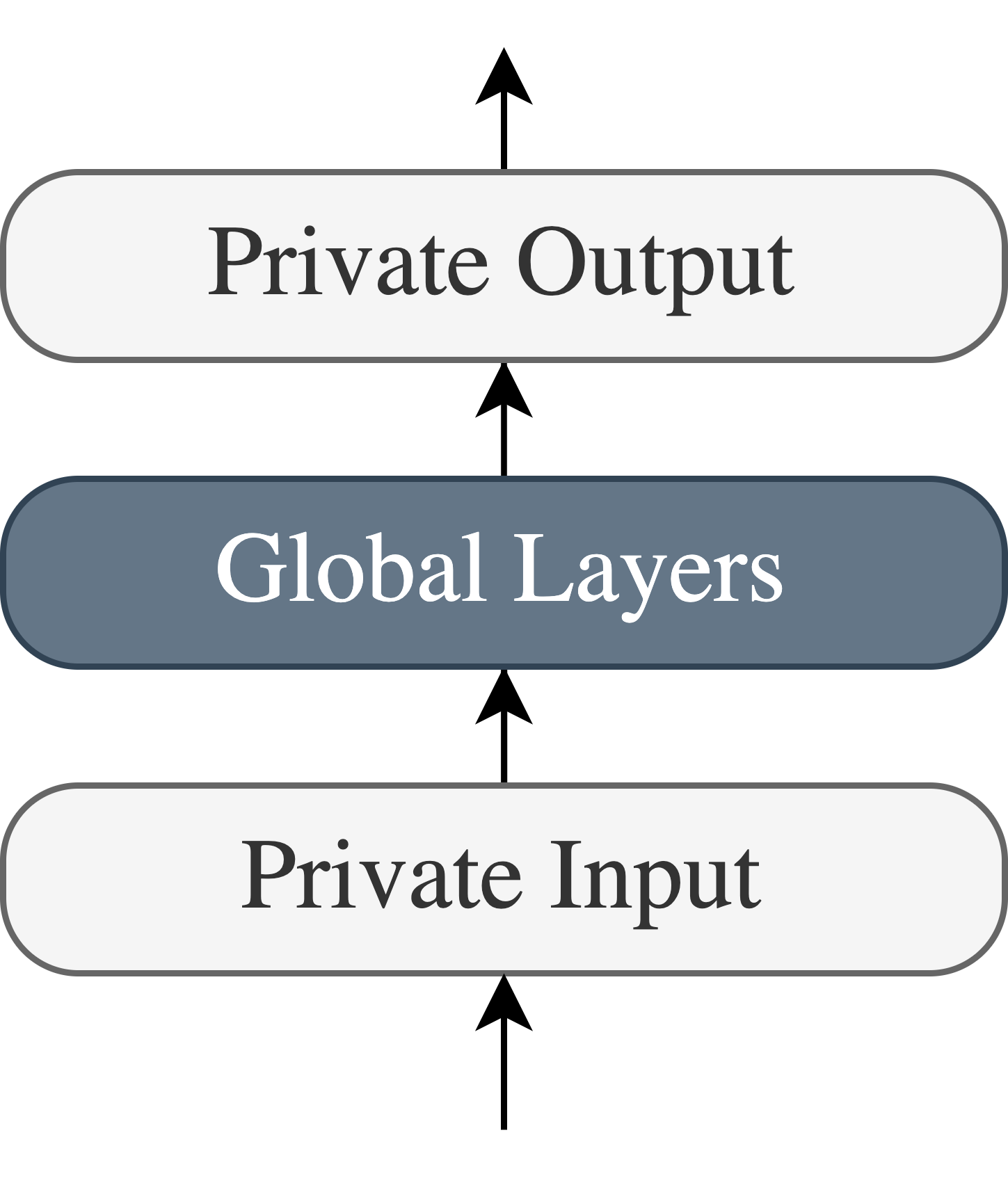}
\caption{GL high-level architecture. 
Each client has a local copy of this model.
Clients use the full model for inference after training is completed. 
During training, only the Global Layers are federated.
The Private Input and Output layers weights remain local to each client and do not get federated.
In general the design of each layer implementation is task dependent.
In Figures \ref{image:model} and \ref{image:submodel} we show the model that we empirically found to work well for both experimental settings, with hyperparameters indicated in section \ref{appendix:hyperparameters}.
}
\label{image:concept}
\end{figure}

The GL model architecture is straightforward conceptually.
Private input and output layers transform local client-specific feature space and dimensions to a global latent space, with consistent dimension across clients such that aggregation such as FedAvg can occur, and project this back to a local client-specific target space.
This is shown in Figure \ref{image:concept}.

Earlier we stated the parameters $w_k = \{u_k, \omega, v_k\}$ from equations \ref{eq:gl} and \ref{eq:gl-sample} are architecturally agnostic and should in general be selected like any other neural-network layers, influenced by task choice, complexity, and the extent of statistical heterogeneity.
For example, a single linear layer with non-linear activation might be appropriate for each of the private and global layers in certain settings.
This point should be explored more in future work.
Optimal layer architectures corresponding to particular non-IID FL settings remains as an open question.

\begin{figure}[p]
\centering
\includegraphics[width=0.9\linewidth]{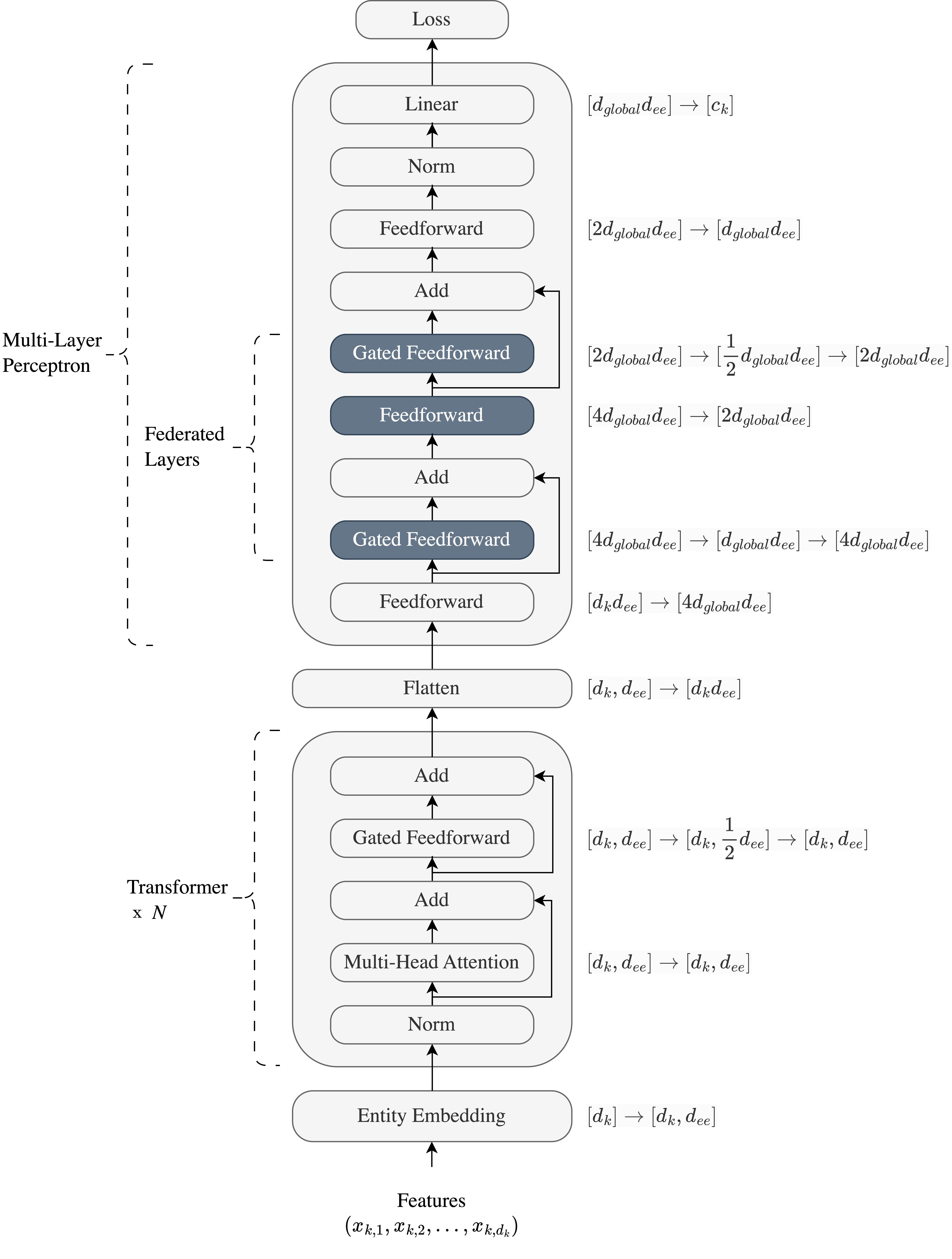}
\caption{GL model architecture. 
Federated layers indicated in darker blue shading; other parameters are private. 
Baseline methods for comparison use an identical architecture, though no parameters are federated for both Local and Centralized settings, and all parameters are federated for the standard FL FedAvg case.
Dimension changes as a result of transformations shown on right, corresponding to embedding sizes we used in our experiments. 
Here, $d_k$ and $c_k$ are the number of features and classes respectively for client $k$, $d_{global}$ is the global embedding dimension and $d_{ee}$ is the Entity Embedding dimension, also common across all clients. 
Implementations for Entity Embedding, Feedforward, and Gated Feedforward layers are shown in Figure \ref{image:submodel}.}
\label{image:model}
\end{figure}

\begin{figure}[h]
\centering
\includegraphics[width=0.95\linewidth]{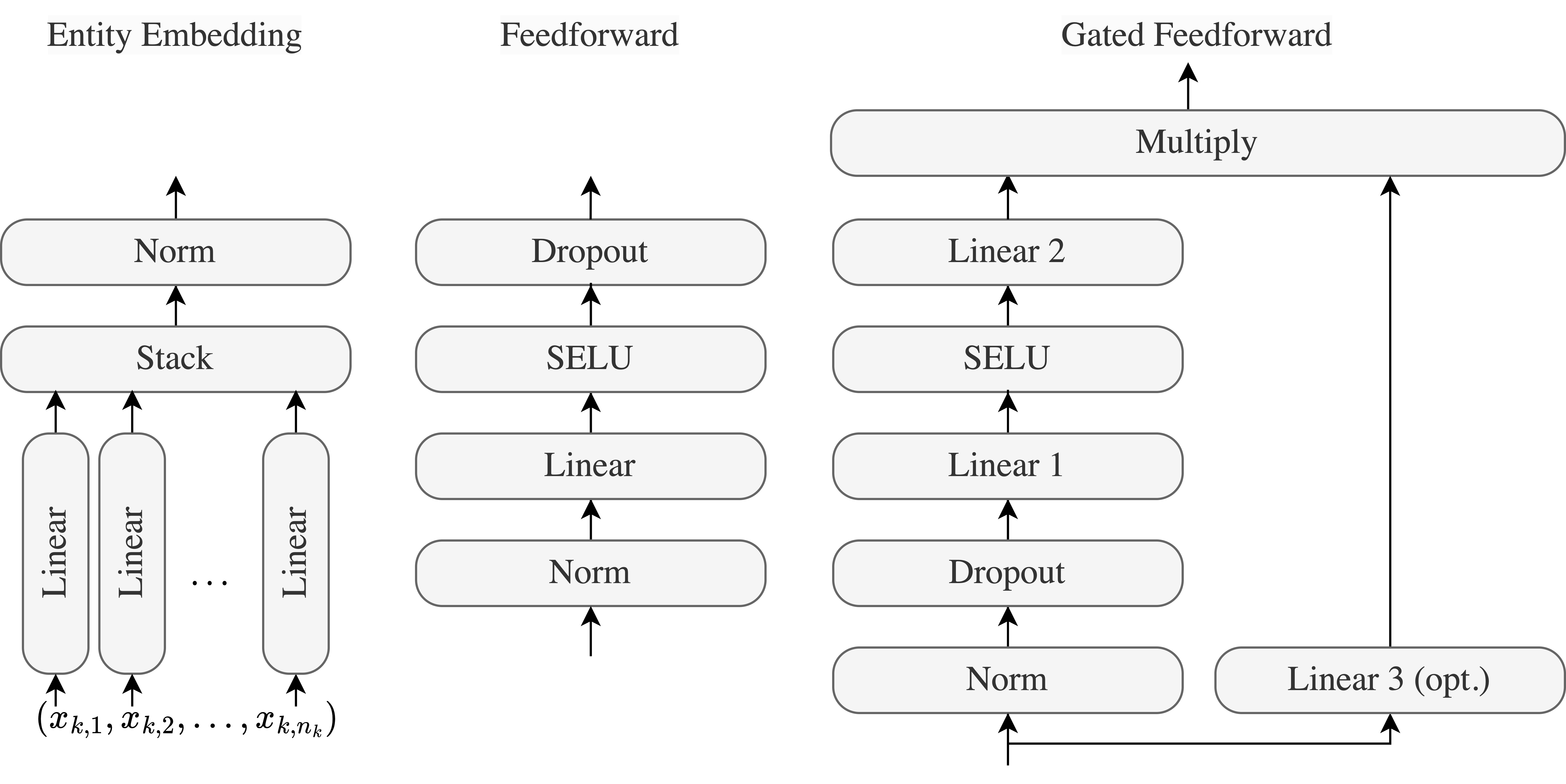}
\caption{(left) Entity Embedding. (centre) Feedforward. (right) Gated Feedforward.}
\label{image:submodel}
\end{figure}

In our experimental settings we found superior empirical performance across all settings, including baselines, with the GL model architecture shown in Figure \ref{image:model}.
The design is influenced by some recent works investigating methods for tabular data, which we reference.
Layer choices and the embedding sizes were selected through hyperparameter optimization with random search cross-validation.
We now describe the main components of this model, each shown in more detail in Figure \ref{image:submodel}.

\subsubsection{Entity Embedding}
\label{appendix:ee}

The main idea of Entity Embedding \citep{guo16} is to map discrete values of a variable to a continuous vector space where relationships between values can be arbitrarily learned.
In Equation \ref{eq:ee} we show this transformation.
The entity embedding dimension, choice of bias for the linear layers and normalization type are hyperparameters.
See Figure \ref{image:submodel}.

Later we describe a preference for ordinal encoding applied to discrete data types.
This is justified as following.
The use of Entity Embedding and its role in learning relationships of objects within a given category suggests that discrete data types should be kept in their ordinal encoding.
One-hot encoding would destroy the class membership category association between discrete object values that Entity Embedding would otherwise use.

\subsubsection{Transformer}

Our use of Transformer \citep{vaswani17} is motivated by Tab-Transformer \citep{huang21}, though \citep{zhang_iisan_21} is similar, and others referenced in Section \ref{model-architecture}.
These works, like ours, use dot-product self-attention to capture complex higher-order feature interactions.
Attention mechanisms allow the model to selectively re-weigh feature contributions to the resulting prediction.
Exploring different attention mechanisms or their uses, for example SAINT \citep{somepalli2021} which proposes \textit{intrasample} attention, is a fruitful area for future work.

A notable work that emphasizes the importance and power of representation learning in the tabular domain is SubTab \citep{ucar21}, which recently showed that tabular self-supervised latent representation learning from feature subsets on images in tabular format can outperform CNN-based SOTA models over a 2D image format.

\subsubsection{Multi-Layer Perceptron}

The original Tab-Transformer specifies a Multi-Layer Perceptron (MLP) implementation consisting of 2 hidden layers and one output layer with sizes $\{4d_{in}, 2d_{in}, d_{c}\}$, where $d_{in}$ is the input size from the previous layer, and $d_c$ is the number of classes.
This works well with, for example, the UCI Adult dataset in the centralized setting, which was originally reported.
However for our non-IID FL experiments we found it necessary to extend this with additional capacity.
The embedding sizes we use are shown on the right side of Figure \ref{image:model}, though their selection in general, beyond our experimental settings, should be treated as hyperparameters.

In GL there are 2 repeating blocks of Feedforward and Gated Feedforward with residual, and an additional Feedforward layer before the linear output layer.
The implementations for Feedforward and Gated Feedforward are described next.
This grants up to 4 layers with consistent dimensionality across clients that we can be federated assuming only the first Feedforward and final output reflect local client dimensionality.
Though it would be more precise to say that in this case 8 linear projections in total are global layer candidates that may be federated.

\subsubsection{Feedforward} 

The Feedforward layer is a normalized linear layer with non-linearity and dropout.
In other words, a standard non-linear feedforward projection.
See Figure \ref{image:submodel}.
In our experiments we found that Batch Normalization, Scaled Exponential Linear Unit (SELU) \citep{klambauer17} activation and no dropout, i.e. set to $0.0$ works well, though again in general these should be treated as hyperparameters.

\subsubsection{Gated Feedforward} 

The Gated Feedforward layer consists of 3 linear projections $w_{gff,1}: \mathbb{R}^{d_{in}} \rightarrow \mathbb{R}^{d_{hidden}}$ and $w_{gff,2}: \mathbb{R}^{d_{hidden}} \rightarrow \mathbb{R}^{d_{in}}$ which project to and from a hidden dimension, generally smaller to encourage learning better representations, and $w_{gff,3}: \mathbb{R}^{d_{in}} \rightarrow \mathbb{R}^{d_{in}}$ which applies multiplicative gating to the former.
A non-linearity is applied after the first projection only, we used SELU in our experiments, and dropout before.
See Figure \ref{image:submodel}.

\subsection{Federated Layers}

We propose that the choice of layers to federate should be treated as any other hyperparameter and selected with cross-validation.
The only restriction is that the dimensionality must be consistent across client parameters so that they may be aggregated.
We found that typical strategies for hyperparameter selection such as random search with cross-validation works well.
Exploring optimal selections of which partial-model parameters to federate within a set of inner consistent-dimension candidates is an important open research question.

\subsubsection{Personalizing Global Invariants}

In Section \ref{learning-objective}, we propose learning a global equipredictive $\omega$ such that: 
$P(Y_k | v_{k} ( \omega (u_{k} (X_k)))) \,\forall k \in K$, 
where $u_{k}$ and $v_{k}$ are local input and output layers, respectively.
To briefly illustrate how a client may compose and personalize the global $\omega$, consider the following re-paramaterization, 
where $\circ$ denotes composition and assuming invertible layers.
\begin{equation} 
\label{eq:gl-reparam}
v_k\circ(\omega\circ u_k) = \underset{\hat{v_k}} {\underbrace{(v_k\circ{v}^{\prime}_k)}} \circ \left( \underset{\hat{\omega}} {\underbrace{(\omega^{\prime\prime}\circ\omega\circ{\omega}^{\prime})}} \circ \underset{\hat{u_k}} {\underbrace{(u^{\prime}_k\circ u_k)}} \right)
\end{equation}
If ${v}^{\prime}_k=\hat{\omega}^{-1}$ then a client $k$ may disregard the global contribution entirely, and similarly for ${u}^{\prime}_k=\hat{\omega}^{-1}$.
On the other hand if ${v}_k={{v}^{\prime}}^{-1}_k$ or ${u}_k={{u}^{\prime}}^{-1}_k$, then it may ignore local contributions.
Finally, either ${v}^{\prime}_k={\omega^{\prime\prime}}^{-1}$ or ${\omega^{\prime}}^{-1}={u}^{\prime}_k$ allows the client to isolate a specific client-invariant transformation of interest $\omega$ while other clients may use the full $\hat{\omega}$.
We leave a full theoretical treatment for future work.

\section{Datasets}
\label{appendix:datasets}

In this section we elaborate on the datasets used in our experiments.
Specifically we describe the construction of 2 sets of datasets that form our 2 experiments.
The first is based on the classic Covertype benchmark tabular dataset which was naturally multi-source at the time of collection.
The second is a collection of 3 multi-source medical datasets independently collected from 3 different hospitals globally.
The datasets may be downloaded from the public repositories given in Table \ref{tab:dataset-sources}.

Justifying this, to the best of our knowledge, there are no prior existing tabular FL benchmark datasets comprising of real-world multi-source clients with natural client feature or class exclusivity.
Only 1/4 related non-IID FL works mentioned in Section \ref{introduction} ran experiments on tabular data \citep{liu_completely_2022}, and they assumed only disjoint but overlapping feature spaces with a shared target space.
Further, their clients were generated from single-source centralized datasets with random feature space and sample partitioning.
This type of construction constrains the resulting expressed heterogeneity as explained in Section \ref{experiment-settings} and therefore is not a realistic representation of what may occur in real-world settings.

\subsection{Covertype}

The original Covertype dataset is tabular format and contains 54 features: 40 one-hot encoded columns representing soil types, 4 one-hot encoded columns representing wilderness area, 3 columns represent morning, midday and afternoon shade, and 7 other numerical columns.
The task is to predict one of 7 classes representing forest cover types.

\begin{table*}[h]
\begin{tabular}{ c | c } 
\toprule
 Feature & Type \\
\hline
elevation & float32  \\
aspect & float32 \\
slope & float32 \\
hhdist & float32 \\
vhdist & float32 \\
hrdist & float32 \\
hfpdist & float32 \\
soil\_type & int \\
climate\_zone & int \\
geological\_zone & int \\
wilderness\_area &  int \\
\hline
\end{tabular}
\centering
\caption{Constructed "Covertype11" dataset features and their data type. Numerical feature names correspond to original dataset feature names.}
\label{tab:cover11-features}
\end{table*}

\subsubsection{Pre-processing}

As motivated in Section \ref{appendix:ee}, we reversed the pre-existing one-hot encoding into ordinal encoding for two features.
The first one was from 40 columns representing soil types which we omit due to length.
From these we constructed an additional 2 features.
Each column names is 4 digits and the original documentation \textit{covtype.info} specifies that the first digit represents the climate zone, and the second digit represents the geological zone.
The second ordinal feature was wilderness area, reversed from 4 one-hot encoded columns with names: \textit{rawah}, \textit{neota}, \textit{comanche}, \textit{poudre}.
All of these ordinal features are integers types.

To summarize, we converted 40 one-hot columns into 1 ordinal column, from this feature generated 2 more, and finally converted 4 columns to 1 ordinal feature we use as a partition key and drop.
We drop the 3 shade columns.
The 7 numerical features are all encoded as float32 but otherwise they remain unchanged.
In total, before partitioning i.e. with a centralized dataset we are left with the 11 features shown in Table \ref{tab:cover11-features}.

After cross-validation and client partitioning, ordinal features were ordinal encoded, numerical features were standardized and normalized and targets were label encoded. 
All encoders were fitted privately and only to the training cross-validation split.
The ordinal encoder default missing value was set to $-1$.
We used Scikit-Learn's implementation \citep{scikit-learn} for all 3 encoders.

\subsubsection{Cross-Validation} 
We re-use the same train/test/validation splits as described in \textit{covtype.info} with shuffling: first 11,340 records used for training data subset, next 3,780 records used for validation data subset, last 565,892 records used for testing data subset.

\subsubsection{Client Partitioning}
The Covertype dataset contains samples from 4 distinct regions in the Roosevelt National Forest of northern Colorado.
We encourage referencing the original \textit{covtype.info} documentation from the UCI repository in Table \ref{tab:dataset-sources} which describes distinct differences between these regions.
Therefore we argue such client partitioning is justified by reverting the centralization that was done after initial data collection.

As a result of partitioning, we encountered different but overlapping classes between clients, shown in Table \ref{tab:cover-labels}.
This is not problematic for local-only training or partially-personalized FL methods client-specific outputs layers such as GL.
Therefore these are label encoded as normal.
However standard FL with full-parameter federation depends on same-sized parameters, so to obtain a comparison result for this case we encode each client with a globally fitted label-encoder.
In other words we force alignment between client label spaces for the standard full-model aggregation case with FedAvg.

\begin{table*}[h]
\begin{tabular}{ c | c | c | c | c } 
\toprule
 Dataset & Num. & Local Classes & Local Label-Encoded & Global Label-Encoded  \\
\hline
Comanche & 6 & [1, 2, 3, 5, 6, 7] & [0, 1, 2, 3, 4, 5] & [0, 1, 2, 4, 5, 6] \\
Neota & 3 & [1, 2, 7] & [0, 1, 2] & [0, 1, 6]\\
Poudre & 4 & [2, 3, 4, 6] & [0, 1, 2, 3] & [1, 2, 3, 5]\\
Rawah  & 4 & [1, 2, 5, 7] & [0, 1, 2, 3] & [0, 1, 4, 6]\\
Centralized  & 7 & [1, 2, 3, 4, 5, 6, 7] & [0, 1, 2, 3, 4, 5, 6] & -\\
\hline
\end{tabular}
\centering
\caption{Class and resulting labels-encoding for Covertype datasets. 
For local-only and GL, local label-encodings are used.
While for standard full-parameter federation, to achieve a consistent output layer dimension necessary for aggregating, we label-encode each client's local classes with a globally fitted label encoder.
The global label encoder has the same classes as the centralized local label-encoded case.
This also ensures we do not unintentionally add concept shift and concept drift for the standard FL setting.}
\label{tab:cover-labels}
\end{table*}

\subsection{Heart Disease}

In this experiment we use 3 distinct datasets that were independently collected at different hospitals globally and at different points in time.
The feature spaces are completely disjoint, with no overlap existing between datasets.
However, the labels incidentally align.
In all datasets, the "0" label corresponds to a patient without heart disease.
Nonetheless this experiment remains a challenge that a traditional full-parameter global model $w$, or even a single centralized model without drastic feature engineering, could not address due to dimensional mismatch across input spaces.
While we did not explore and compare with possible alternatives such as aligning with missing values, or padding, and this would be valuable future work, these methods are expected to encounter challenges due to the significant, approximately threefold, increase in feature vector dimension they would cause.

\begin{table*}[p]
\begin{tabular}{ c | c } 
\toprule
 Feature & Type \\
\hline
age & int  \\
sex & int \\
cp & int \\
trestbps & int \\
chol & int \\
fbs & int \\
restecg & float32 \\
thalach & float32 \\
exang & int \\
oldpeak & float32 \\
slope &  int \\
ca &  int \\
thal &  float32 \\
\hline
\end{tabular}
\centering
\caption{"Cleveland" from UCI Heart Disease, dataset features and their data type.}
\label{tab:cleveland-features}
\end{table*}

\begin{table*}[p]
\begin{tabular}{ c | c } 
\toprule
 Feature & Type \\
\hline
sbp & int  \\
tobacco & float32 \\
ldl & float32 \\
adiposity & float32 \\
famhist & int \\
typea & int \\
obesity & float32 \\
alcohol & float32 \\
age & int \\
\hline
\end{tabular}
\centering
\caption{South Africa UCI Heart Disease, dataset features and their data type.}
\label{tab:saheart-features}
\end{table*}

\begin{table*}[p]
\begin{tabular}{ c | c } 
\toprule
 Feature & Type \\
\hline
age & float32  \\
anaemia & int \\
creatinine\_phosphokinase & int \\
diabetes & int \\
ejection\_fraction & int \\
high\_blood\_pressure & int \\
platelets & float32 \\
serum\_creatinine & float32 \\
serum\_sodium & int \\
sex & int \\
smoking &  int \\
time &  int \\
\hline
\end{tabular}
\centering
\caption{"Faisalabad" from UCI Heart Failure, dataset features and their data type.}
\label{tab:faisalabad-features}
\end{table*}

\subsubsection{UCI Heart Disease, Cleveland}

The UCI Heart Disease dataset \citep{detrano_international_1989}, and in particular the Cleveland dataset, contains 303 patient case studies of coronary heart and artery disease diagnoses collected from the Cleveland Clinic Foundation in the United States between May 1981 and September 1984 \citep{marateb_noninvasive_2015}.
In this task 14 clinical features, shown in Table \ref{tab:cleveland-features} are used to predict a binary label where $0$ indicates no presence and $1$ indicates presence of heart disease.
Out of the 303 patients, 164 are labeled no presence.
Note that in the UCI repository there are 2 files: \textit{cleveland.data} and \textit{processed.cleveland.data}.
The former is described as "corrupted" in the \textit{WARNING} file in the UCI repository and we thus use the latter.

Besides Cleveland, there are an additional 3 hospital sites: Long Beach VA, Switzerland, and Hungarian, with 200, 123, and 294 instances respectively.
We found that inclusion of these smaller datasets as clients would penalize performance for the others and so we focused only on the Cleveland dataset as typically done and described in the original \textit{heart-disease.names} documentation in the UCI repository.
Similarly, while there are 74 features available we focused on the 14 that are typically used, as well as considering only the binary label version.

\subsubsection{South Africa Heart Disease}

The South Africa Heart Disease dataset is originally from a 1988 retrospective study \citep{rossouw_coronary_1983} that contains 462 samples of 9 features from male individuals in high-risk heart disease regions in the Western Cape of South Africa.
The goal was to explore prevalence of heart disease in conjunction with high-risk features such as smoking and alcohol consumption in these areas.
The features are described in Table \ref{tab:saheart-features}.
Out of the 462 samples, 160 are patients with coronary heart disease, and the remaining 302 are controls.
This binary classification task uses $0$ to represent absence, and $1$ to indicate presence of coronary heart disease.

\subsubsection{UCI Heart Failure, Faisalabad}

The UCI Heart Failure dataset \citep{chicco20} was originally collected \citep{ahmad17} during April–December 2015 at the Faisalabad Institute of Cardiology and the Allied Hospital in Faisalabad, Pakistan.
It contains 299 patient cases of heart failure, specifically left ventricular systolic dysfunction, with ages ranging from 40-95 and 194 men compared with 105 women.
12 features in total, shown in Table \ref{tab:faisalabad-features} are used to predict a binary target where $1$ indicates a death event, and $0$ means survival.
We refer to the original studies for more information.

\subsubsection{Pre-processing}

No pre-processing was done besides standardization and normalization of numerical features, and ordinal encoding of discrete features with $-1$ for missing values.
With one exception.
The FedAvg baseline with standard FL cannot support disjoint feature spaces as stated in Section \ref{baseline-methods}.
Therefore to obtain a result for this case we applied Pandas \citep{reback2020pandas} \textit{DataFrame.Align} with inputed $0$'s for the padded features belonging to other clients.
This results in a consistent feature space which allows full-parameter federation.
As we did not create a centralized dataset with such feature space alignment we do not report a centralized baseline for the Heart Disease experiment.

Similar to the Covertype experiment, all encoders were fitted privately and only to the training cross-validation split, and again using Scikit-Learn's implementation.

\subsubsection{Cross-Validation} 

Cross-validation is achieved first by splitting the original dataset into 70\% training and 30\% testing, and then another 30\% of the split training data used as a validation hold-out set.
We again used Scikit-Learn's \textit{traintestsplit} implementation with random seed so that each experiment replication generates a new seeded train/test/validation set with shuffling.

\subsubsection{Client Partitioning}

For the Heart Disease experiment no partitioning was needed as each client is a single-source dataset.

\section{Experiments}
\label{appendix:experiments}

So far we have described the clients that undergoing FL in each experiment, as well as their model architectures.
At the time of developing our experiments, there were no published methods to handle any client input or output space heterogeneity, at all.
This motivated the choice of baseline methods we described in Section \ref{experiment-settings}.

We developed an experimentation module that would replicate the same model training scenario, with random seed, for each comparison method, given a single launch command.
During training, statistics are recorded against the train set after each epoch, as well as a validation set that runs by default each epoch as well.
After training, the separate holdout test set is evaluated to produce timeseries metrics for each seeded run, as well as their summary across all runs, and finally a summary over all methods.
Runs are fully deterministic and reproducible.

In both experiments we use AdamW optimizer.
Covertype is multi-class, so we use Cross-Entropy loss.
Heart Disease is binary class, so we use Binary Cross-Entropy.
Optimizer, loss and models are implemented with PyTorch.

The number of parameters in each model setting is given in Table \ref{tab:cover-num-param} for Covertype and Table \ref{tab:heart-num-param} for Heart Disease.

Each client model is initialized randomly with the same random seed.

\begin{table*}[h]
\begin{tabular}{ c | c c c c c c} 
\toprule
\multicolumn{1}{c}{} & \multicolumn{2}{c}{Local} & \multicolumn{2}{c}{FedAvg} & \multicolumn{2}{c}{GL} \\
\hline
 & Num. Param. & Federated & Num. Param. & Federated & Num. Param. & Federated \\
\hline
Comanche & 915954 & 915954 & 916131 & 916131 & 915954 & 717640 \\
Neota & 915954 & 915954 & 916131 & 916131 & 915423 & 717640 \\
Poudre & 915954 & 915954 & 916131 & 916131 & 915600 & 717640 \\
Rawah & 915954 & 915954 & 916131 & 916131 & 915600 & 717640 \\
\hline
\end{tabular}
\centering
\caption{Number of Model Parameters in Covertype experiment. "Num. Param" corresponds to all model parameters; "Federated" corresponds to the parameters that are aggregated.}
\label{tab:cover-num-param}
\end{table*}

\begin{table*}[h]
\begin{tabular}{ c | c c c c c c} 
\toprule
\multicolumn{1}{c}{} & \multicolumn{2}{c}{Local} & \multicolumn{2}{c}{FedAvg} & \multicolumn{2}{c}{GL} \\
\hline
 & Num. Param. & Federated & Num. Param. & Federated & Num. Param. & Federated \\
\hline
Cleveland & 448339 & 448339 & 597415 & 597415 & 448339 & 298688 \\
South Africa & 415211 & 415211 & 597415 & 597415 & 415211 & 298688 \\
Faisalabad & 440057 & 440057 & 597415 & 597415 & 440057 & 298688 \\
\hline
\end{tabular}
\centering
\caption{Number of Model Parameters in Heart Disease experiment. "Num. Param" corresponds to all model parameters; "Federated" corresponds to the parameters that are aggregated.}
\label{tab:heart-num-param}
\end{table*}

\subsection{Hyperparameters}
\label{appendix:hyperparameters}

The hyperparameters used in both experiments are provided in Table \ref{tab:common-hyp-param}.

\begin{table*}[h]
\begin{tabular}{ c | c } 
\toprule
 Hyperparameter & Value \\
\hline
Entity Embedding Dimension & 16 \\
Learning Rate & 0.001  \\
Weight Decay &  0.0001 \\
Dropout &  0.0 \\
Transformer Heads & 8 \\
Transformer Blocks & 6 \\
$a_k$ & 1.0 \\
Global Layer Update Rate & 1.0 \\
\hline
\end{tabular}
\centering
\caption{Hyperparameters used in both experiments.}
\label{tab:common-hyp-param}
\end{table*}

Hyperparameters particular to Covertype experiment is in Table \ref{tab:cover-hyp-param} and for Heart-Disease in Table \ref{tab:heart-hyp-param}.

\begin{table*}[h]
\begin{tabular}{ c | c } 
\toprule
 Hyperparameter & Value \\
\hline
Epochs & 105 \\
Batches & 10 \\
Random Seeds & 8:28 \\
$d_{global}$ & 8 \\
\hline
\end{tabular}
\centering
\caption{Hyperparameters used in Covertype experiment.}
\label{tab:cover-hyp-param}
\end{table*}

\begin{table*}[h]
\begin{tabular}{ c | c } 
\toprule
 Hyperparameter & Value \\
\hline
Epochs & 10 \\
Batches & 15 \\
Random Seeds & 0:101 \\
$d_{global}$ & $d_k=11$ \\
Loss & Binary Cross-Entropy \\
\hline
\end{tabular}
\centering
\caption{Hyperparameters used in Heart Disease experiment.}
\label{tab:heart-hyp-param}
\end{table*}

\subsection{Numerical Results}
\label{appendix:numerical-results}

In Tables \ref{tab:numerical-results-auroc}, \ref{tab:numerical-results-acc}, and \ref{tab:numerical-results-auprc} we show numerical results obtained in the experimental settings we have described.
Mean, standard deviation and 95\% confidence results are provided across 21 and 101 random seeds for Covertype and Heart Disease respectively.

We observe a clear and nearly universal increase in performance and lower variance across AUROC, Accuracy (Balanced Accuracy for Heart Disease), and AUPRC with GL over the baseline methods.

It is interesting to observe that some clients report higher than the centralized baseline in the Covertype experiment.
This suggests the personalized global invariants may in fact be more robust than what centralized training alone can learn.
This should be explored more in future work, particularly in context of the idea that FL is a natural setting for robust OOD learning \citep{francis21}.
On the other hand, for the Heart Disease experiment there is no baseline method applicable due to disjoint feature spaces.
We leave for future work to compare against the case of a composite union of feature spaces as previously mentioned.

\subsection{Reproducibility}
\label{reproducibility}

All experimental code is provided in the following repository: \url{https://github.com/transferFL/gl/}.

Documentation to prepare the datasets and run the experiments is provided in the project root README.md.

The four datasets used in the experiments are available on public repositories, provided in Table \ref{tab:dataset-sources}.
These are already available in our code repository, however they may also be re-downloaded and placed into folders according to the README.md.
Table \ref{tab:our-dataset-urls} links to the folders in our code repository containing the exact dataset files we used.

\begin{table*}[h]
\begin{tabular}{| c | c |} 
\toprule
 Original Dataset & Source URL \\
\hline
Covertype & https://archive.ics.uci.edu/ml/datasets/covertype \\
Heart Disease (Cleveland) & https://archive.ics.uci.edu/ml/datasets/Heart+Disease  \\
South Africa Heart Disease &  http://statweb.stanford.edu/\~tibs/ElemStatLearn/data.html \\
Heart Failure (Faisalabad) &  https://archive.ics.uci.edu/ml/datasets/Heart+failure+clinical+records \\
\hline
\end{tabular}
\centering
\caption{Original dataset source URLs.}
\label{tab:dataset-sources}
\end{table*}

\begin{table*}[h!]
\begin{tabular}{| c | c |} 
\toprule
 Experiment & Source URL \\
\hline
Covertype & \url{https://github.com/transferFL/gl/tree/main/cover/data}  \\
Heart Disease & \url{https://github.com/transferFL/gl/tree/main/heart/data} \\
\hline
\end{tabular}
\centering
\caption{URLs to datasets used in our experiments.}
\label{tab:our-dataset-urls}
\end{table*}

The numerical results in Tables \ref{tab:numerical-results}, \ref{tab:numerical-results-auroc}, \ref{tab:numerical-results-acc}, and \ref{tab:numerical-results-auprc} will be replicated exactly when the random seeds specified in Section \ref{appendix:hyperparameters} are used with a CPU device, Torch 2.1.0 with Python 3.11.2.

Each experiment may be launched with a single command provided in the README.md.

\newpage

\begin{table*}[p]
\begin{tabular}{| c c | c c c | c c c } 
\toprule
\multicolumn{2}{|c}{} & \multicolumn{3}{|c|}{AUROC (\%)}  \\
\hline
 Experiment & Dataset & Local & FedAvg & \textbf{GL} \\
\hline
\multirow{3}{*}{\shortstack{Covertype \\ (n=21)}} 
& Comanche & $88.29 \pm 8.36 \; (3.58)$ & $79.58 \pm 7.89 \; (3.37)$ & $\mathbf{92.09 \pm 0.80 \; (0.34)}$  \\
& Neota & $79.44 \pm 9.46 \; (4.05)$ & $60.92 \pm 2.03 \; (0.87)$ & $\mathbf{88.85 \pm 1.90 \; (0.81)}$ \\
& Poudre & $86.74 \pm 1.63 \; (0.70)$ & $63.89 \pm 3.33 \; (1.42)$ & $\mathbf{88.88 \pm 1.53 \; (0.65)}$ \\
& Rawah  & $88.86 \pm 1.53 \; (0.65)$ & $69.82 \pm 3.76 \; (1.61)$ & $\mathbf{91.43 \pm 0.65 \; (0.28)}$ \\
& Centralized  & $88.93 \pm 10.71 \; (4.58)$ & $-$ & $-$ \\
\hline
\multirow{3}{*}{\shortstack{Heart Disease \\ (n=101)}}
& Cleveland & $88.57 \pm 3.11 \; (0.61)$ & $\mathbf{89.67 \pm 2.61 \; (0.51)}$ & $89.59 \pm 2.89 \; (0.56)$   \\
& South Africa & $74.49 \pm 3.82 \; (0.75)$ & $71.13 \pm 4.44 \; (0.87)$ & $\mathbf{76.02 \pm 3.70 \; (0.72)}$  \\
& Faisalabad & $85.64 \pm 4.09 \; (0.80)$ & $83.68 \pm 4.59 \; (0.90)$ & $\mathbf{86.77 \pm 3.80 \; (0.74)}$  \\
\hline
\end{tabular}
\centering
\caption{Mean and standard deviation cross-validation result reported across 21, 101 random seeds for Covertype, Heart Disease respectively. 95\% confidence interval shown in parenthesis.}
\label{tab:numerical-results-auroc}
\end{table*}

\begin{table*}[p]
\begin{tabular}{| c c | c c c | c c c } 
\toprule
\multicolumn{2}{|c}{} & \multicolumn{3}{|c|}{(Balanced) Accuracy$^1$ (\%)} \\
\hline
 Experiment & Dataset & Local & FedAvg & \textbf{GL} \\
\hline
\multirow{3}{*}{\shortstack{Covertype \\ (n=21)}} 
& Comanche & $80.82 \pm 6.95 \; (2.97)$ & $80.46 \pm 4.98 \; (2.13)$ & $\mathbf{83.81 \pm 0.45 \; (0.19)}$  \\
& Neota & $76.43 \pm 6.67 \; (2.85)$ & $72.26 \pm 6.55 \; (2.80)$ & $\mathbf{81.13 \pm 1.90 \; (0.81)}$ \\
& Poudre & $72.57 \pm 1.20 \; (0.51)$ & $72.72 \pm 1.51 \; (0.65)$ & $\mathbf{74.21 \pm 0.94 \; (0.40)}$ \\
& Rawah  & $84.25 \pm 0.65 \; (0.28)$ & $83.65 \pm 0.68 \; (0.29)$ & $\mathbf{85.13 \pm 0.44 \; (0.19)}$ \\
& Centralized  & $79.70 \pm 15.55 \; (6.63)$ & $-$ & $-$ \\
\hline
\multirow{3}{*}{\shortstack{Heart Disease \\ (n=101)}}
& Cleveland  & $80.23 \pm 3.98 \; (0.78)$ & $78.06 \pm 4.27 \; (0.83)$ & $\mathbf{81.18 \pm 3.70 \; (0.72)}$ \\
& South Africa  & $62.43 \pm 3.94 \; (0.77)$ & $56.15 \pm 5.20 \; (1.01)$ & $\mathbf{63.22 \pm 4.22 \; (0.82)}$ \\
& Faisalabad  & $74.57 \pm 5.13 \; (1.00)$ & $67.66 \pm 6.54 \; (1.28)$ & $\mathbf{75.15 \pm 4.79 \; (0.93)}$ \\
\hline
\end{tabular}
\centering
\caption{Mean and standard deviation cross-validation result reported across 21, 101 random seeds for Covertype, Heart Disease respectively
($^1$) Heart Disease binary classification reports balanced accuracy. 95\% confidence interval shown in parenthesis.}
\label{tab:numerical-results-acc}
\end{table*}

\begin{table*}[p]
\begin{tabular}{| c c | c c c | c c c } 
\toprule
\multicolumn{2}{|c}{} & \multicolumn{3}{|c|}{AUPRC (\%)} \\
\hline
 Experiment & Dataset & Local & FedAvg & \textbf{GL} \\
\hline
\multirow{3}{*}{\shortstack{Covertype \\ (n=21)}} 
& Comanche & $57.82 \pm 10.76 \; (4.60)$ & $45.37 \pm 9.20 \; (3.93)$ & $\mathbf{65.52 \pm 2.56 \; (1.09)}$  \\
& Neota & $70.23 \pm 12.43 \; (5.31)$ & $27.88 \pm 3.53 \; (1.51)$ & $\mathbf{83.11 \pm 3.61 \; (1.54)}$ \\
& Poudre & $67.85 \pm 3.53 \; (1.51)$ & $26.67 \pm 2.78 \; (1.19)$ & $\mathbf{71.83 \pm 2.94 \; (1.26)}$ \\
& Rawah  & $64.94 \pm 3.65 \; (1.56)$ & $31.98 \pm 5.15 \; (2.20)$ & $\mathbf{69.90 \pm 3.19 \; (1.36)}$ \\
& Centralized  & $53.09 \pm 13.70 \; (5.86)$ & $-$ & $-$ \\
\hline
\multirow{3}{*}{\shortstack{Heart Disease \\ (n=101)}}
& Cleveland  & $89.04 \pm 3.59 \; (0.70)$ & $89.57 \pm 3.13 \; (0.61)$ & $\mathbf{90.29 \pm 3.08 \; (0.60)}$ \\
& South Africa  & $59.71 \pm 5.66 \; (1.10)$ & $54.88 \pm 6.13 \; (1.20)$ & $\mathbf{61.86 \pm 5.91 \; (1.15)}$ \\
& Faisalabad  & $74.34 \pm 7.38 \; (1.44)$ & $71.67 \pm 7.60 \; (1.48)$ & $\mathbf{76.41 \pm 6.76 \; (1.32)}$ \\
\hline
\end{tabular}
\centering
\caption{Mean and standard deviation cross-validation result reported across 21, 101 random seeds for Covertype, Heart Disease respectively. 95\% confidence interval shown in parenthesis.}
\label{tab:numerical-results-auprc}
\end{table*}

\end{document}